\documentclass[conference]{IEEEtran}
\usepackage{graphicx}
\usepackage{subcaption}
\usepackage{url}
\usepackage{multirow}
\usepackage{array}
\usepackage{amsmath}
\ifCLASSINFOpdf
\else
\fi

\usepackage{tikz}
\begin{document}
%
\title{Firearm Detection via Convolutional Neural Networks: Comparing a Semantic Segmentation Model Against End-to-End Solutions}


\makeatletter
\newcommand{\linebreakand}{%
  \end{@IEEEauthorhalign}
  \hfill\mbox{}\par
  \mbox{}\hfill\begin{@IEEEauthorhalign}
}
\makeatother

\author{\IEEEauthorblockN{Alexander Egiazarov}
\IEEEauthorblockA{\textit{Digital Security Group} \\
\textit{University of Oslo}\\
Oslo, Norway  \\
alexegiazarov@gmail.com}
\and
\IEEEauthorblockN{Fabio Massimo Zennaro}
\IEEEauthorblockA{\textit{Digital Security Group} \\
\textit{University of Oslo}\\
Oslo, Norway \\
fabiomz@ifi.uio.no}
\linebreakand
\IEEEauthorblockN{Vasileios Mavroeidis}
\IEEEauthorblockA{\textit{Digital Security Group} \\
\textit{University of Oslo}\\
Oslo, Norway \\
vasileim@ifi.uio.no}
}


%


\maketitle

\begin{abstract}
Threat detection of weapons and aggressive behavior from live video can be used for rapid detection and prevention of potentially deadly incidents such as terrorism, general criminal offences, or even domestic violence. One way for achieving this is through the use of artificial intelligence and, in particular, machine learning for image analysis. 
In this paper we conduct a comparison between a traditional monolithic end-to-end deep learning model and a previously proposed model based on an ensemble of simpler neural networks detecting fire-weapons via semantic segmentation. We evaluated both models from different points of view, including accuracy, computational and data complexity, flexibility and reliability. Our results show that a semantic segmentation model provides considerable amount of flexibility and resilience in the low data environment compared to classical deep model models, although its configuration and tuning presents a challenge in achieving the same levels of accuracy as an end-to-end model.

\end{abstract}


%
\IEEEpeerreviewmaketitle

\section{Introduction}

Threat detection from live video feeds of firearms, knives, and aggressive behavior can be used in preventing or rapidly detecting and mitigating potentially deadly incidents such as terrorism, general criminal offenses, or even domestic violence. One way for achieving this is the use of artificial intelligence and, in particular, machine learning for image analysis to detect weapons that, in many cases, can also be partially concealed, thus making their discovery a difficult task.

According to the report "Global Study on Homicide" published by the United Nations Office on Drugs and Crime, 54\% of homicides in 2017 involved firearms, accounting for 238,804 victims. That proves that firearms are a preferred instrument for committing a crime. In addition, firearms may be used in mass shootings thus resulting in multiple loss of lives in a single incident, such as in the case of a terrorist act.

In previous work, we presented an approach for firearm detection that makes use of an ensemble of Semantic Convolutional Neural Networks \cite{egiazarov2020firearm}. This approach decomposes a task, such as the detection of a firearm, into a set of smaller tasks, such as the detection of individual component parts of the firearm. We argued that this approach has computational and practical advantages compared to the traditional single monolithic approach, such as requiring less computational resources for training the smaller models and the ability to train the individual component part models in parallel. The results of our previous work demonstrated that the individual networks achieved satisfactory accuracy after being trained on a limited set of data. An important strength of this approach is that the final system relies not only on the performance of the individual networks but also on the ensembling of the results of all networks.

In this paper, we put to rigorous test our hypotheses about the strengths and weaknesses of an approach based on semantic segmentation. We perform a series of experimental simulation to assess the accuracy, the flexibility, and the robustness of our model against the end-to-end model based on a single deep network. Our conclusions clearly delineate the advantages, as well as the significant limitations, of our solution.

\section{Background}

In this section, we introduce the problem of weapon detection and we explain how it can be expressed as a machine learning problem. We then recall the principle of semantic segmentation or decomposition, and we describe two approaches to the problem of weapon detection: an approach based on a single deep neural network, and the approach based on an ensemble of simpler semantic neural networks. 

\subsection{Weapon Detection and Segmentation}
The majority of the weapon detection solutions employ  classical machine learning methods where the object is classified or localized by common computer vision techniques or using monolithic architectures \cite{tiwari2015computer}. Despite the popularity of the semantic segmentation approach \cite{Arbelaez2011,MostajabiYS14,Fulkerson2009,Chen2014} very few strides were made in the implementation of weapon detection systems based on semantic decomposition, primarily because it requires a radically different approach to design of the deep learning models and unique datasets. 


\subsection{Machine Learning}
Machine learning provides a set of methods and techniques for inferring patterns and relations among data.
More formally, in the \emph{supervised learning} setting, we are given a collection of $N$ data samples $x_i$ each one with its own label $y_i$; a supervised learning algorithm allows us to learn a general function $f: x_i \mapsto y_i$ that maps data samples onto their respective labels \cite{bishop2006pattern}. In the specific case of weapon detection and segmentation, the set of samples $x_i$ will correspond to images, while the set of labels $y_i$ will correspond either to binary values denoting the presence of a weapon in an image or to a box surrounding a weapon within an image.

A versatile algorithm for learning a mapping $f$ is offered by \emph{neural networks}, layered graphical models that can approximate any function (to an arbitrary degree given a sufficiently wide or deep architecture) \cite{bishop2006pattern}.
In the case of images, a special family of neural networks that have been proven to be particularly successful are \emph{deep convolutional neural networks} (CNN) \cite{Krizhevsky2012}. Deep CNNs are neural networks that use convolutional windows to analyze an image and rely on several layers for processing. Thanks to their priors and their complex architecture, deep CNNs are able to learn to discriminate images with high accuracy. The main drawback of this solution lies in its sample requirements and its computational complexity. In order to train a deep CNN, it is necessary to collect a large amount of data and rely on considerable computational power to process this data.

\subsection{Semantic Segmentation}
Semantic segmentation is the general engineering principle of decomposing a single complex task along semantic lines in order to define a set of simpler problems. 
In the specific case of weapon detection, the application of this principle translates into the decomposition of the hard task of detecting whole weapons into a set of easier image detection problems. Since a whole weapon constitutes a complex object in terms of shape, texture, and orientation within a picture, we considered the possibility of decomposing the problem of detecting a single weapon into the problem of detecting some of its visually prominent component parts, such as barrel, stock, magazine, and receiver. Each one of these component parts has a simpler shape, a consistent texture, and a higher degree of orientational invariance than the whole rifle, thus constituting a simpler detection problem.


\subsection{Firearm Detection Based on a Single Neural Network}
\label{single-neural-network}
The basic approach to weapon detection applies the standard paradigm of deep CNNs that has been proven to be successful in image recognition.
This paradigm is based on the implementation of a single deep CNN trained end-to-end with labeled data. After proper training, such a network is expected to be able to accurately detect weapons within images. Deep CNN have the ability to learn complex functions allowing classification of objects and they constitute the state of the art in image detection and segmentation. The drawbacks of CNNs are the immediate consequences of their size and complexity. First, in order to fit a deep model described by a high number of parameters, it is necessary to collect a large amount of data. This may be expensive or challenging, as it is in the case of images of weapons. A second challenge is due to the structural complexity of a deep CNN. As a versatile function fitter, a deep CNN is defined by a large set of hyper-parameters. Properly choosing or exploring a subset of all the possible combinations of hyper-parameters is a non-trivial task. A third challenge derives from those mentioned above. As the amount of data and the complexity of a network increase, so is the computational cost for training the network.






\subsection{Firearm Detection Based on Semantic Decomposition}
We have proposed an alternative approach to the problem of firearm detection based on the principle of semantic segmentation \cite{egiazarov2020firearm}. Instead of designing a single deep CNN for discriminating a whole weapon within an image, we implemented a set of shallower CNNs, each one tasked with the simpler objective of recognizing a single component part of a weapon. The final decision on the presence or absence of a weapon is then achieved by aggregating the outputs of the smaller networks.
This solution allows us to tackle the main drawbacks of a monolithic CNN described in Section \ref{single-neural-network}. In particular, shallower CNNs demand less data and computational power for training, as well as having a smaller space of hyper-parameters. Moreover, we suggest that relying on the outputs of multiple independent networks may make our solution more reliable in situations where weapons are partly obfuscated within an image. Finally, we may be able to achieve a more robust decision by aggregating multiple outputs, as proposed by the theory of ensemble models \cite{rokach2010pattern}. The main weakness of our solution lies in the decomposition itself. In our model, each network learns exclusively to recognize a single component part of a weapon independently from the remaining. A solution based on a single deep CNN may model higher-order correlations between the parts so that detecting one component may help to detect other components.

\section{Problem Statement \label{sec:PS}}
This work experimentally evaluates and compares our semantic segmentation approach to weapon detection against the standard approach based on a deep CNN.

Earlier results \cite{egiazarov2020firearm} demonstrated that our approach for weapon detection and segmentation achieves a reasonable performance. In \cite{egiazarov2020firearm}, it was shown that four small CNNs could be successfully trained for detecting individual component parts of a specific weapon (AR15). The output of the networks could be merged to generate heatmaps and to decide whether a weapon is present or absent.

In this paper, we provide a more careful examination of the proposed semantic segmentation model by comparing it against a single network model resembling more closely the state of the art. In particular, we are interested in studying and comparing the performance of the two models in a regime with limited amount of data and limited computational power. We define a set of tasks intended to provide a fair comparison between our model and the standard model with different capacities. 
Our experiments are not meant to compare our solution directly with the state of the art for deciding which model achieves the highest performance. Instead, we carry out a comparison between scaled-down implementations of our solution and state-of-the-art solutions in order to evaluate the strengths and the weaknesses of our model, in particular in terms of \emph{accuracy} (measured in terms of statistical accuracy), \emph{computational and data cost} (evaluated in terms of architecture depth), \emph{flexibility} (expressed in the compositionality of the outputs of our individual networks) and \emph{reliability} (expressed in the tuning of false positives and false negatives).

Before presenting our simulations, we first describe our models and datasets.

\section{Models}
In this section we describe the models and the architectures we will use in our evaluations.

\subsection{Semantic Segmentation Model}
The aim of the semantic segmentation model is to decompose the hard task of detecting a whole weapon in a set of simpler task aimed at detecting only specific component parts of a weapon.
In order to successfully recognize an AR15 rifle, we identified four main component parts: stock, magazine, barrel and receiver (see Figure \ref{fig:components}). We selected these components as they are the most visually distinct parts of a firearm. 

\begin{figure}[h]
\centering
\caption{Main components of AR-15 style rifle. \label{fig:components}}
\includegraphics[width=\columnwidth]{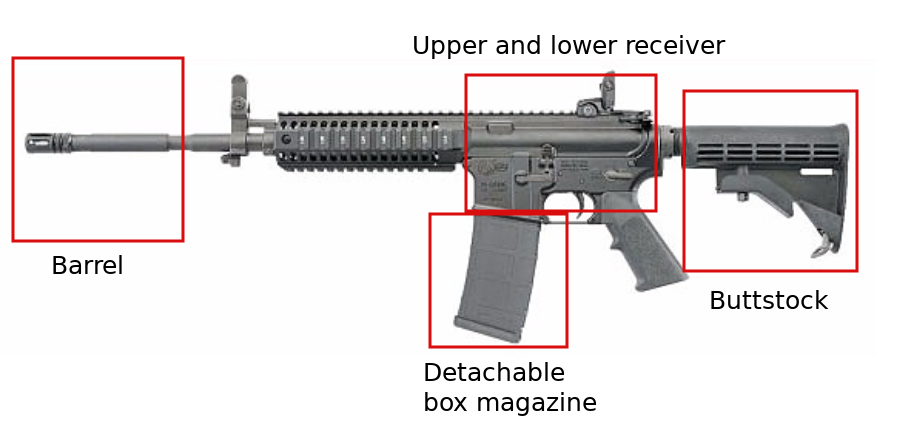}
\end{figure}


The semantic segmentation model has been designed as a multi-layered system; for an illustration, refer to Figure \ref{fig:SSmodel}.

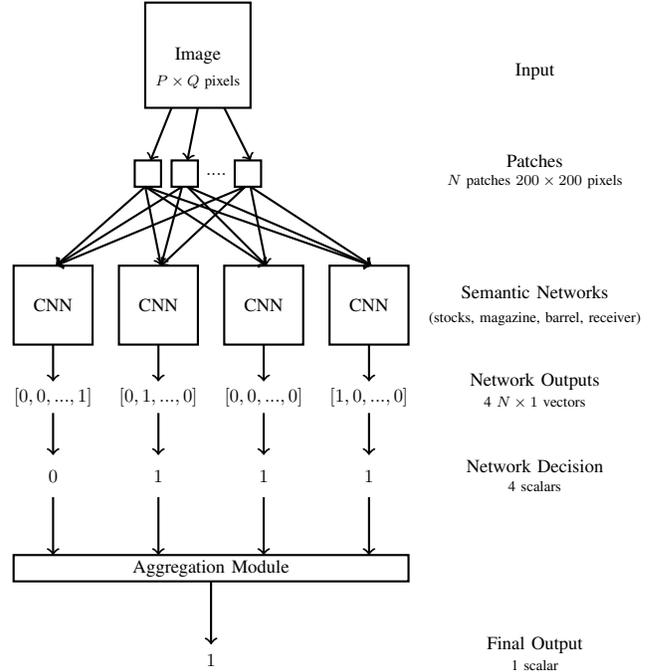
\begin{figure}
\caption{Semantic Segmentation Model \label{fig:SSmodel}}
\begin{tikzpicture}[thick,scale=0.7, every node/.style={scale=0.7}]

\draw  (0,0) rectangle (2,2) node[pos=.5] {Image};
\node[] at (1,0.5) {\footnotesize{$P \times Q$ pixels}};

\node[] at (7.4,0.7) {Input};

\draw  (-.2,-1) rectangle (.3,-1.5);
\draw  (.5,-1) rectangle (1,-1.5);
\node[] at (1.35,-1.25) {{$....$}};
\draw  (1.7,-1) rectangle (2.2,-1.5);

\path[,->] (0.5,0) edge (0.1,-1) ;
\path[,->] (1,0) edge (0.8,-1) ;
\path[,->] (1.5,0) edge (2,-1) ;

\node[] at (7.4,-1) {Patches};
\node[] at (7.4,-1.4) {\footnotesize{$N$ patches $200 \times 200$ pixels}};

\draw  (-2.5,-3) rectangle (-1,-4.5) node[pos=.5] {CNN};
\draw  (-.5,-3) rectangle (1,-4.5) node[pos=.5] {CNN};
\draw  (1.5,-3) rectangle (3,-4.5) node[pos=.5] {CNN};
\draw  (3.5,-3) rectangle (5,-4.5) node[pos=.5] {CNN};

\path[,->] (0,-1.5) edge (-1.7,-3) ;
\path[,->] (0,-1.5) edge (0.3,-3) ;
\path[,->] (0,-1.5) edge (2.3,-3) ;
\path[,->] (0,-1.5) edge (4.3,-3) ;

\path[,->] (.7,-1.5) edge (-1.7,-3) ;
\path[,->] (.7,-1.5) edge (0.3,-3) ;
\path[,->] (.8,-1.5) edge (2.3,-3) ;
\path[,->] (.8,-1.5) edge (4.3,-3) ;

\path[,->] (1.9,-1.5) edge (-1.7,-3) ;
\path[,->] (1.9,-1.5) edge (0.3,-3) ;
\path[,->] (2,-1.5) edge (2.3,-3) ;
\path[,->] (2,-1.5) edge (4.3,-3) ;

\node[] at (7.4,-3.5) {Semantic Networks};
\node[] at (7.4,-4) {\footnotesize{(stocks, magazine, barrel, receiver)}};

\node[] at (-1.75,-5.5) {$[0, 0, ..., 1]$};
\node[] at (0.25,-5.5) {$[0, 1, ..., 0]$};
\node[] at (2.25,-5.5) {$[0, 0, ..., 0]$};
\node[] at (4.25,-5.5) {$[1, 0, ..., 0]$};

\path[,->] (-1.75,-4.5) edge (-1.75,-5.2) ;
\path[,->] (0.25,-4.5) edge (0.25,-5.2) ;
\path[,->] (2.25,-4.5) edge (2.25,-5.2) ;
\path[,->] (4.25,-4.5) edge (4.25,-5.2) ;

\node[] at (7.4,-5.2) {Network Outputs};
\node[] at (7.4,-5.6) {\footnotesize{4 $N\times 1$ vectors}};

\node[] at (-1.75,-7) {$0$};
\node[] at (0.25,-7) {$1$};
\node[] at (2.25,-7) {$1$};
\node[] at (4.25,-7) {$1$};

\path[,->] (-1.75,-5.8) edge (-1.75,-6.6) ;
\path[,->] (0.25,-5.8) edge (0.25,-6.6) ;
\path[,->] (2.25,-5.8) edge (2.25,-6.6) ;
\path[,->] (4.25,-5.8) edge (4.25,-6.6) ;

\node[] at (7.4,-6.8) {Network Decision};
\node[] at (7.4,-7.2) {\footnotesize{4 scalars}};

\path[,->] (-1.75,-7.4) edge (-1.75,-8.5) ;
\path[,->] (0.25,-7.4) edge (0.25,-8.5) ;
\path[,->] (2.25,-7.4) edge (2.25,-8.5) ;
\path[,->] (4.25,-7.4) edge (4.25,-8.5) ;

\draw  (-2.5,-8.5) rectangle (5,-9) node[pos=.5] {Aggregation Module};

\path[,->] (1.25,-9) edge (1.25,-10.2) ;

\node[] at (1.25,-10.5) {$1$};

\node[] at (7.4,-10.2) {Final Output};
\node[] at (7.4,-10.6) {\footnotesize{1 scalar}};

\end{tikzpicture}
\end{figure}

At the \emph{input level}, the model receives an image of arbitrary dimension.
On the following \emph{patches level} the input image is divided into patches. In order to deal with images with different sizes and ratios, a sliding window algorithm is used to extract patches. The size and the step of the sliding window is set in a user-independent fashion as a ratio between the sides of the image. After extraction, each patch is rescaled to $200 \times 200$ pixels. 
On the \emph{semantic networks level} patches are fed into the four component CNNs. These networks are defined in a modular way, following \cite{egiazarov2020firearm}; each CNN is constituted by $M_{sem}$ convolutional layers aimed at performing feature extraction, and $N_{sem}$ dense layers carrying out the final classification. In the convolutional section, we use layers containing 32 or 64 filters with default stride of 1x1, with ReLU activation functions, and 2x2 max-pooling \cite{egiazarov2020firearm}.
In the dense section, we use fully connected layers; we use a ReLU activation function, except for the last layer where we rely on the softmax function to compute the output. Moreover, in the second-to-last layer, we use dropout \cite{Srivastava2013a} with a probability $p=0.5$ for regularization \cite{egiazarov2020firearm}. Given the relatively small architectures, all the networks are trained independently, in parallel, on their respective labelled data.
On the \emph{network outputs level} we collect the output of each individual network. Each CNN produces an array of arbitrary length, made up of binary values; each value denotes the presence or the absence of a specific component in each patch. 
On the \emph{network decision level} the vector of binary values outputted by each network is aggregated into a final decision, representing the evaluation on whether a weapon component was present in the original image. Aggregating a binary vector of arbitrary length into a single value may be accomplished using different algorithms, from a voting mechanism to processing these vectors using a dedicated module, such as a recurrent neural network able to manage arbitrary-length inputs. We rely on validated thresholds: we consider the outputs on all the patches, and we use thresholds to decide whether few positive network outputs on isolated patches constitute a false positive, or whether a concentrated set of positive networks outputs flagged the presence of a weapon component.
Lastly, on the \emph{final output level} the binary decisions of the four network are aggregated in the overall decision of the model about the presence or absence of a weapon. A basic solution would be to simply use a decision algorithm which counts the outputs from each of the four network modules as 25\% probability of presence. However, this algorithm may be improved by a more sophisticated decision algorithms that process the output of the individual networks. We illustrate different voting and weighting mechanisms aimed at maximizing the final accuracy of the model. 



\subsection{Single Network Model}
As a comparison, we instantiate a solution shaped on the state of the art for weapon detection. We define a deep CNN trained on whole images in order to learn to flag the presence of a weapon inside the image. For illustration, refer to the model in Figure \ref{fig:DNNmodel}.

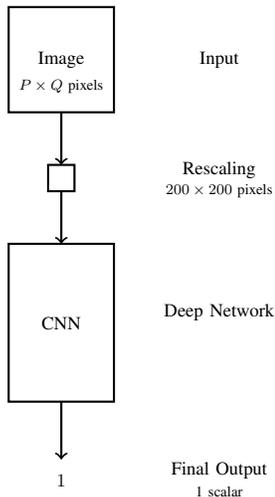
\begin{figure}
\begin{center}
\begin{tikzpicture}[thick,scale=0.7, every node/.style={scale=0.7}]

\draw  (0,0) rectangle (2,2) node[pos=.5] {Image};
\node[] at (1,0.5) {\footnotesize{$P \times Q$ pixels}};

\node[] at (4,1) {Input};

\draw  (.75,-1) rectangle (1.25,-1.5);

\path[,->] (1,0) edge (1,-1) ;

\node[] at (4,-1.1) {Rescaling};
\node[] at (4,-1.5) {\footnotesize{$200 \times 200$ pixels}};

\path[,->] (1,-1.5) edge (1,-2.5) ;
\draw  (0,-2.5) rectangle (2,-5.5) node[pos=.5] {CNN};

\node[] at (4,-3.8) {Deep Network};

\node[] at (1,-7) {$1$};

\path[,->] (1,-5.5) edge (1,-6.6) ;

\node[] at (4,-6.8) {Final Output};
\node[] at (4,-7.2) {\footnotesize{1 scalar}};

\end{tikzpicture}
\end{center}
\caption{Single Network Model \label{fig:DNNmodel}}

\end{figure}

Like the semantic segmentation model the individual network model receives at the \emph{input level} an image of arbitrary dimension.
On the following \emph{rescaling level} the input image is rescaled to a fixed dimension.
On the \emph{deep network level} the image is forwarded to a deep CNN which is made up of $M_{single}$ convolutional layers, and $N_{single}$ dense layers, with the same hyper-parameters described above for the CNNs in the semantic segmentation model.
Finally, on the \emph{final ouput level} we obtain the output of the deep CNN in the form of a (uncalibrated) probability of the firearm being present in the picture.

As discussed in Section \ref{sec:PS}, we do not aim at implementing a cutting edge architecture for the sake of achieving the best possible performance. In other words, we are not interested in pushing the number and the width of layers  $M_{single}$ and $N_{single}$ as high as possible. Instead, we will pay attention to the ratio between the number of layers in the semantic segmentation model ($M_{sem}$ and $N_{sem}$) and in the deep CNN ($M_{single}$ and $N_{single}$). This will allow us to evaluate the relative performance of the two models with respect to the computational power or data availability necessary to train deeper models.

\section{Datasets}

In this section we discuss the generation and the preparation of the data for our models.

\subsection{Data for the Semantic Segmentation Model}
The architecture of the semantic segmentation model requires the definition of a custom dataset for each one of the four CNNs included in the system. Specifically, each CNN has to be trained on a proper dataset that contains positive samples (images of the weapon component that the network is supposed to detect) and negative samples (random images not containing the weapon component that the network is supposed to detect). 

To create the datasets for the semantic segmentation model we assembled a total of $4500$ images from the public domain. We chose to use publicly available images, instead of synthetically created ones, due to to the higher variation of details and combination of components that are naturally present in the sourced dataset.
We visually inspected the original set of images to verify its quality and removed any sample that did not portray the actual chosen firearm model (e.g., obvious toy replicas, other firearm models) or depicted images with non-related content. We then extracted $2500$ positive patches and $2500$ negative patches for each of the four CNNs. All patches were resized to 200$\times$200 pixels size. Positive-labelled patches contain the specific component part for each network, while negative-labelled patches contain random images (including background details, clothing, people and other random objects available in the starting set of images) and samples of other component parts that the network is not supposed to detect. Notice that enriching the negative dataset with component parts that the network should not detect is essential to prevent the network from learning to detect just the color or the texture of rifle parts, instead of the actual component part. See Figure \ref{fig:barrel_samples} for examples of positive-labelled and negative-labelled samples.

\begin{figure}[h]
\centering
\caption{Data samples. On the left, a general purpose negative sample; on the right, a positive sample for the barrel network and a potential negative sample for the receiver, magazine and buttstock network. \label{fig:barrel_samples}}
\includegraphics[width=0.4\columnwidth]{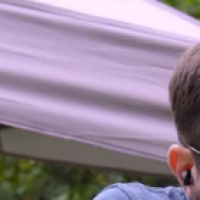}
\includegraphics[width=0.4\columnwidth]{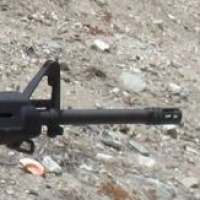}
\end{figure}

Each dataset is partitioned into a training, validation and testing subset with the respective proportions of 80\%, 16\% and 4\%, thus yielding 2000 training samples, 400 validation samples and 100 testing samples.

Each training dataset is then augmented via random modification of the samples (such as, rotation, offset from center and scale changes).
By adding 3 modified images in addition to the original sample, each training dataset was enlarged to 8000 samples. This procedure provides a bigger variety of data and, by applying augmentation after partitioning our data, we guarantee that the same samples with and without modification will not appear in the training and test dataset. Inclusion of modified images changed the training, validation and testing set proportions to 95\%, 5\% and 1\% respectively.

At the end, each CNN is trained and evaluated on a dataset made up of a training set $\mathcal{D}^{tr}_{sem}$ of $16000$ samples, a validation set $\mathcal{D}^{val}_{sem}$ of $800$ samples, and a test set $\mathcal{D}^{te}_{sem}$ of $200$ samples, all evenly divided in positive and negative samples. See Table \ref{tab:image-partition-table} for a summary of the data and its partitioning.

\begin{table}[h]
\caption{Dataset for each CNN in the semantic segmentation model.}
\label{tab:image-partition-table}
\begin{center}
\begin{tabular}{lcccc}
Positive patches                                       & \textbf{Training}         & \textbf{Validation}      & \textbf{Testing}         & \textbf{Total}            \\ \hline
\multicolumn{1}{|l|}{\textbf{Initial partitioning}} & \multicolumn{1}{c|}{2000} & \multicolumn{1}{c|}{400} & \multicolumn{1}{c|}{100} & \multicolumn{1}{c|}{2500} \\ \hline
\multicolumn{1}{|l|}{\textbf{After augmentation}}  & \multicolumn{1}{c|}{8000} & \multicolumn{1}{c|}{400} & \multicolumn{1}{c|}{100} & \multicolumn{1}{c|}{8500} \\ \hline
Negative patches                                       & \textbf{Training}         & \textbf{Validation}      & \textbf{Testing}         & \textbf{Total}            \\ \hline
\multicolumn{1}{|l|}{\textbf{Initial partitioning}} & \multicolumn{1}{c|}{2000} & \multicolumn{1}{c|}{400} & \multicolumn{1}{c|}{100} & \multicolumn{1}{c|}{2500} \\ \hline
\multicolumn{1}{|l|}{\textbf{After augmentation}}  & \multicolumn{1}{c|}{8000} & \multicolumn{1}{c|}{400} & \multicolumn{1}{c|}{100} & \multicolumn{1}{c|}{8500} \\ \hline
\end{tabular}
\end{center}
\end{table}

\subsection{Data for the Single Network Model}
For the single neural network model we collected a new dataset. This decision is due to the fact that the deep CNN is meant to be trained to detect a whole weapon within an image and therefore it can not be trained on the patches used to train the four component-specific CNNs.
Positive-labelled sample are extracted from the original dataset of $4500$ public domain images. We selected $3500$ images containing the entirety of the AR15 rifle. As before, this set of $3500$ images is partitioned into a training dataset of 3000 samples, a validation dataset of 400 samples and a test dataset of 100 samples. The training dataset is further augmented using the same method used for the semantic segmentation dataset, thus producing a final training dataset of 8000 samples. Negative-labelled samples are extracted from the Indoor Scene Recognition\footnote{\url{http://web.mit.edu/torralba/www/indoor.html}} dataset. This dataset contains varied and realistic images of indoor environments, which may resemble the places where an automatic weapon detection system may be deployed. We randomly sub-selected 8000 images for training, 400 for validation, and 100 for testing. We carefully selected from different categories, and, given the abundance of data, we did not perform any augmentation.  
At the end, the CNN in the single network model is trained and evaluated on a dataset made up of a training set $\mathcal{D}^{tr}_{single}$ of $16000$ samples, a validation set $\mathcal{D}^{val}_{single}$ of $800$ samples, and a test set $\mathcal{D}^{te}_{single}$ of $200$ samples, all evenly divided in positive and negative samples. See Table \ref{tab:image-partition-table2} for a summary of the data and its partitioning.

\begin{table}[h]
\caption{Dataset for the CNN in the single network model.}
\label{tab:image-partition-table2}
\begin{center}
\begin{tabular}{lcccc}
Positive patches                                       & \textbf{Training}         & \textbf{Validation}      & \textbf{Testing}         & \textbf{Total}            \\ \hline
\multicolumn{1}{|l|}{\textbf{Initial partitioning}} & \multicolumn{1}{c|}{3000} & \multicolumn{1}{c|}{400} & \multicolumn{1}{c|}{100} & \multicolumn{1}{c|}{3500} \\ \hline
\multicolumn{1}{|l|}{\textbf{After augmentation}}  & \multicolumn{1}{c|}{8000} & \multicolumn{1}{c|}{400} & \multicolumn{1}{c|}{100} & \multicolumn{1}{c|}{8500} \\ \hline
Negative patches                                       & \textbf{Training}         & \textbf{Validation}      & \textbf{Testing}         & \textbf{Total}            \\ \hline
\multicolumn{1}{|l|}{\textbf{Initial partitioning}} & \multicolumn{1}{c|}{8000} & \multicolumn{1}{c|}{400} & \multicolumn{1}{c|}{100} & \multicolumn{1}{c|}{8500} \\ \hline
\end{tabular}
\end{center}
\end{table}

\section{Evaluation}

In this section, we present the evaluation of the two models we have described above, the \emph{semantic decomposition model} and the \emph{single network model}. Our simulations are meant to compare the two solutions, highlighting the different performances of each module, showing the degrees of freedom in aggregating the individual networks in the semantic decomposition model, and contrasting the results in the two models when trained on very limited sets of data.

\subsection{Simulation 1: Evaluating the CNNs}
In this simulation, we train independently all the CNNs of our two models, using different settings for their hyperparameters. We perform model selection and choose the optimal architecture for each of the CNN we implemented. For the semantic segmentation model, this simulation runs up to the \emph{network outputs level} of Figure \ref{fig:SSmodel}.

\paragraph{Protocol}
In these simulations we consider different architectures for the individual component-specific CNNs and for the single deep CNN. In particular we vary the number of the convolutional layers ($M_{sem}, M_{single}$) and the number of dense layers ($N_{sem}, N_{single}$) in the set $\left\{3,4,5\right\}$. These values were chosen to include the basic setting for the semantic decomposition model described in \cite{egiazarov2020firearm}, and to allow for the exploration of larger models with higher capacity, within a modest computational budget for training.

In total, we considered $5$ possible architectures ($(M=3,N=3), (M=3,N=4), (M=4,N=3), (M=4,N=4), (M=5,N=5)$), $10$ models ($5$ semantic segmentation models and $5$ single network models), leading to the training, validation and testing of $25$ networks ($4$ CNNs for each semantic segmentation model and $1$ CNN for each single network model).

Training is performed for 15 iterations, and we use the validation dataset to perform early stopping and select the weight configuration returning the best accuracy on the validation dataset.
Notice that in this simulation each network is trained and evaluated independently. In the semantic segmentation model, the four component CNNs are evaluated at the \emph{network output level} (see Figure \ref{fig:SSmodel}), therefore ignoring for the moment the \emph{network decision level} and the \emph{final output level}.

\paragraph{Results}
Table \ref{tab:perfromanceBest} reports the best architectures found in our model selection process (more results are available in the Appendix). For each network we report the architecture expressed as the number of convolutional layers, $M_{single}$ or $M_{sem}$, and dense layers, $N_{single}$ or $N_{sem}$. We also report the early stopping epoch and the associated accuracy on the validation dataset. The final performance of the network is expressed in terms of true positives and true negatives over the test dataset.

\begin{table}[hbt!]
\caption{Best architecture (\emph{arch}) for each network in the two models, epoch of early stopping (\emph{epoch}), performance on the validation dataset at the early stopping epoch ((\emph{acc (val)})) and true positives (\emph{TP (test)}) and true negatives (\emph{TN (test)}) on the test set. \label{tab:perfromanceBest}}
\begin{center}
\begin{tabular}{|l|c|c|c|c|c|}
\hline
                   & \textbf{Arch} & \textbf{Epoch} & \textbf{Acc (val)} & \textbf{TP (test)} & \textbf{TN (test)} \\ \hline
\textbf{Full AR}   & 4x4             & 15    & 94,6\%                 & 95\%                 & 90\%                 \\ \hline
\textbf{Barrels}   & 4x4             & 6     & 95,2\%                 & 99\%                 & 95\%                 \\ \hline
\textbf{Magazines} & 4x3                   & 12    & 94,1\%                 & 96\%                 & 93\%                 \\ \hline
\textbf{Receivers} & 5x5                   & 8     & 97,7\%                 & 99\%                 & 90\%                 \\ \hline
\textbf{Stocks}    & 5x5                   & 11    & 94,8\%                 & 98\%                 & 90\%                 \\ \hline
\end{tabular}
\end{center}
\end{table}

All the networks achieve a good performance in their training, using architectures of similar complexity. The single network model achieves the best validation performance with $4$ convolutional layers and $4$ dense layers. Its performance on this dedicated task is, in general, inferior to the individual component networks; this result is understandable as the single network model tackles here a more complex task (recognizing a whole weapon) compared to the networks in the semantic segmentation model (recognizing component parts).  

\paragraph{Discussion}
This simulation allowed us to select the optimal architectural hyper-parameters for our two models. It is likely that given more training data, larger architecture with stricter regularization, and more computational power, these results could be improved upon. Within the limits we selected in terms of computational powers and number of layers, the hyper-parameters we found constitute the optimal solutions for our models and we will use these hyper-parameters in the following simulations.

\subsection{Simulation 2: Tuning of the Network Decisions}

In the previous simulation we evaluated the performance of individual component networks in detecting weapon components within individual patches. We now estimate the threshold parameters that would allow us to combine the outputs over each patch into a final decision. This simulation runs only for the semantic segmentation model at the \emph{network decision level} of Figure \ref{fig:SSmodel}.

\paragraph{Protocol} In this simulation we use the optimal CNNs that we have already trained. These networks have been trained so far only to classify patches of images, and decide whether each of them contains the specific component part they were trained on. In order to process a whole image, we need to aggregate the outputs of each CNN over several patches. 

Given an image $x$, each network records its network outputs for each patch extracted from the image. The juxtaposition of these network outputs for overlapping patches provides us with a detection heatmap $H(x)$ over the image $x$. 

Using the validation data, we can estimate data-defined thresholds from the heatmaps $H(x)$ in order to return a network decision. We estimate a \emph{positive threshold} $\theta_p$ as the average of the maximum value of the heatmap for the positive images:
\[
\theta_p = E \left[ \max_{x \in \mathcal{P}}  H(x) \right],
\]
where $\mathcal{P}$ is the set of positive images. Similarly we evaluate a \emph{negative threshold} $\theta_n$  as the average of the maximum value of the heatmap for the negative images:
\[
\theta_n = E \left[ \max_{x \in \mathcal{N}}  H(x) \right],
\]
where $\mathcal{N}$ is the set of negative images. Finally we also define an intermediate threshold $\theta_i$ which is estimated on the combined positive set P and negative set N:

\[
\theta_i = E \left[ \max_{x \in (\mathcal{P} \cup \mathcal{N})}  H(x) \right].
\]
Notice that, in our case, since the validation dataset is balanced, we have $\theta_i = \frac{\theta_p + \theta_n}{2}$ because of the linearity of the expectation.

Thus, given an image $x$, each network will process all the patches, compute the heatmap $H(X)$ over the image, evaluate the mean heatmap $E \left[ H(X) \right]$, and compare it against one of the learned thresholds $\theta$. The output will be a positive decision if $E \left[ H(X) \right] \geq \theta$. The gap between thresholds may be used to define an uncertainty region which may require the intervention of a human supervisor in the loop. However, in this simulation, we will simply return a negative decision if $E \left[ H(X) \right] < \theta$.

We compute the thresholds of each component network using the $400$ positive and $400$ negative images in the validation dataset for the single network model $\mathcal{D}^{val}_{single}$. 
Notice that in this simulation we do no explicitly evaluate the accuracy of the each component network because the images in the validation dataset $\mathcal{D}^{val}_{single}$ and test dataset $\mathcal{D}^{te}_{single}$ of the single network model are labelled in terms of presence of a whole weapon, and they lack labels about the presence of individual components; indeed, it is not unusual that in an image containing a weapon, one of the four components may be occluded or hidden; such an image, while being a positive instance of a weapon, would be a negative instance for the occluded component. An overall evaluation of the semantic segmentation model in terms of accuracy is therefore postponed to the next section; here we estimate possible thresholds $\theta$ for the \emph{network decision level}.


\paragraph{Results}
Table \ref{tab:theta_vals} reports the thresholds computed on the validation data. As expected $\theta_p > \theta_n$ for all the networks; this makes sense as we would expect positive samples containing instances of weapon to raise detection in more patches. However, in the magnitude of these thresholds we can observe that certain parts may be easier to detect than others; in particular, the closeness of the two thresholds $\theta_p, \theta_n$ for the magazine network may point to the fact that correctly discriminate the presence or absence of magazine may be more difficult that other components.

\begin{table}[hbt!]
\centering
\caption{Threshold values for each component network}
\label{tab:theta_vals}
\begin{tabular}{|l|l|l|l|}
\hline
\textbf{}          & \textbf{$\theta_p$} & \textbf{$\theta_n$} & \textbf{$\theta_i$} \\ \hline
\textbf{Barrels}   & 32.385              & 10.407              & 21.396              \\ \hline
\textbf{Magazines} & 4.345               & 3.462               & 3.903               \\ \hline
\textbf{Receivers} & 24.822              & 3.915               & 14.368              \\ \hline
\textbf{Stocks}    & 45.437              & 6.397               & 25.917              \\ \hline
\end{tabular}
\end{table}

\paragraph{Discussion}
This simulation allowed us to compute the threshold parameters for our component networks which will allow us to compute the overall decision of a component network. Moreover, the same computation of these thresholds has provided us with further insight on the weapon detection problem, highlighting that some weapon component discrimination problems may be harder than others.

\subsection{Simulation 3: Comparing the Semantic Segmentation Model and the Single Network Model}

Building on the previous simulation that allowed us to compute a single decision for each network, in this simulation we finally evaluate the overall performance of the semantic segmentation model against the single network model. We consider different aggregation protocols to merge the decisions of the individual networks, and we compare the accuracy of their final decision against the accuracy of the single network module. This simulation runs at the \emph{final output level} of Figure \ref{fig:SSmodel}.

\paragraph{Protocol}
We run our simulations using the optimal hyper-parameters found in the previous simulations. However, instead of testing the two models independently on their respective datasets as we did in Simulation 1, we contrast their results on an identical dataset. 

A key challenge in this experiment is how to guarantee that the performances of the two models are compared in a fair way. 
First of all, notice that both our models are trained on similar positive samples coming from a common original dataset of $4500$ public domain images; we thus assume that both models are provided with similar training information about the object to detect; although negative samples may differ, we expect the training data for positive samples not to be skewed or manipulated as to provide an advantage to any of the two models.
Next, we need to guarantee that the measure of performance is equitable. For this to be the case, we need to evaluate the performance of the two models on the same test cases. Thus given a test image, the outputs of the two models can be compared in a consistent way. 
 
This comparison is then \emph{fair with respect to the data} (both models learned from sets of data derived from a common source). However, it may be argued whether our comparison is fair with the respect to data processing choices, such as the way outputs in the semantic segmentation model are aggregated or how images are rescaled in the single network models; such choices are specific to each one of the two models, and it is therefore hard to guarantee any sort of fairness with respect to them; we think that the best approach would be to consider these choices as hyper-parameters of the two models and investigate how performances would change when varying these additional hyper-parameters; we will not carry out this further investigation in this paper; instead, we will present our conclusion conditional to our assumptions, and leave further investigation for future work.
At the end, we opted to use the test dataset $\mathcal{D}^{te}_{single}$ we prepared for the single network model. This may provide a small edge to the single network model that was trained on data coming from the same distribution, and it will constitute a realistic out-of-distribution challenge for the semantic segmentation model.

For the aggregation of the four decisions of the individual component networks, we start implementing simple voting rules: \emph{strict majority rule} (final output is positive if at least three out of four networks return a positive decision), \emph{weak majority rule} (final output is positive if at least two out of four networks return a positive decision), \emph{unanimity rule} (final output is positive iff all the four networks return a positive decision) and \emph{veto rule} (final output is positive if at least one out of four networks return a positive decision).
We also consider the possibility of having a weighted vote, in which the weight of each individual component network is scaled with respect to its accuracy; we use the normalized validation accuracy of each network to set the weights. This approach allows to underestimate the decisions of weak networks, and boost the decision of networks performing over average.

At the end, we measure the performances of each model in terms of accuracy.

\paragraph{Results}
Table \ref{tab:results-aggregation-new} shows the accuracy of the semantic segmentation model at the \emph{final output level}, as a function of the different voting rules and the different possible thresholds $\theta$ used at the \emph{network decision level}.

In general, we observe that there seems to be no optimal $\theta$ for all the aggregation rules. On the contrary, we can observe a correlation between the magnitude of $\theta$ and how stringent a rule is. This makes sense: loose rules (like having 1 network out of 4 detecting a component part to flag a detection) may take advantage of a higher $\theta$ threshold to prevent too many false positives; on the opposite, strict rules (like requiring all 4 out of 4 networks detecting the respective component part to flag a detection) may operate better with a lower $\theta$ that would avoid too many false negatives. The type of rule and the magnitude of $\theta$ may then be jointly set or optimized as hyperparameters in order to control the trade-off between precision and recall.

The weighted vote has also been tested. However, due to the almost-uniform performance of the component networks (see Table \ref{tab:perfromanceBest}), the weights were very close to be uniform and we did not observe any significant difference in accuracy. 
We still hold, though, that in case of sufficiently different performances, the weighted vote rule may have a positive impact on the overall results.

For a direct comparison with the single network models, these accuracies should be contrasted against the results reported for the \emph{Full AR} architecture in Table \ref{tab:perfromanceBest}. The comparison shows that the single network model easily outperforms the semantic segmentation model. On one side, this may be due to the better fit between training and test data in the case of the single network model and to its ability of modelling correlations between component part; on the other side, the versatility of the semantic segmentation model, presenting the opportunity to tune thresholds and aggregation rules, offers more control to the designer, but also constitute a further challenge in the optimization process.

\begin{table}[hbt!]
\centering
\caption{Final accuracy of the semantic segmentation model as a function of the threshold parameter $\theta$.}
\label{tab:results-aggregation-new}
\small\addtolength{\tabcolsep}{-4pt}
\begin{tabular}{|l|l|l|l|l|}
\hline
\textbf{Rule}       & $\theta = 0$                                                                  & $\theta_n$                                                                   & $\theta_i$                                                                    & $\theta_p$                                                                   \\ \hline
\textbf{1 out of 4} & \begin{tabular}[c]{@{}l@{}}TP = 100\%\\ TN = 27\%\\ Tot = 63.5\%\end{tabular} & \begin{tabular}[c]{@{}l@{}}TP = 99\%\\ TN = 49\%\\ Tot = 74\%\end{tabular}   & \begin{tabular}[c]{@{}l@{}}TN = 91\%\\ TP = 66\%\\ Tot = 78.5\%\end{tabular}  & \begin{tabular}[c]{@{}l@{}}TP = 79\%\\ TN = 77\%\\ Tot = 78\%\end{tabular}   \\ \hline
\textbf{2 out of 4} & \begin{tabular}[c]{@{}l@{}}TP = 94\%\\ TN = 64\%\\ Tot = 79\%\end{tabular}    & \begin{tabular}[c]{@{}l@{}}TP = 86\%\\ TN = 81\%\\ Tot = 83.5\%\end{tabular} & \begin{tabular}[c]{@{}l@{}}TP = 61\%\\ TN = 90\%\\ Tot = 75.5\%\end{tabular}  & \begin{tabular}[c]{@{}l@{}}TP = 39\%\\ TN =95\%\\ Tot = 67\%\end{tabular}    \\ \hline
\textbf{3 out of 4} & \begin{tabular}[c]{@{}l@{}}TP = 60\%\\ TN = 89\%\\ Tot = 74.5\%\end{tabular}  & \begin{tabular}[c]{@{}l@{}}TP = 44\%\\ TN = 95\%\\ Tot = 69.5\%\end{tabular} & \begin{tabular}[c]{@{}l@{}}TP = 25\%\\ TN = 100\%\\ Tot = 62.5\%\end{tabular} & \begin{tabular}[c]{@{}l@{}}TP = 10\%\\ TN = 100\%\\ Tot = 55\%\end{tabular}  \\ \hline
\textbf{4 out of 4} & \begin{tabular}[c]{@{}l@{}}TP = 16\%\\ TN = 98\%\\ Tot = 57\%\end{tabular}    & \begin{tabular}[c]{@{}l@{}}TP = 9\%\\ TN = 100\%\\ Tot = 54.5\%\end{tabular} & \begin{tabular}[c]{@{}l@{}}TP = 3\%\\ TN = 100\%\\ Tot  = 51.5\%\end{tabular} & \begin{tabular}[c]{@{}l@{}}TP = 1\%\\ TN = 100\%\\ Tot = 50.5\%\end{tabular} \\ \hline
\end{tabular}
\end{table}

\paragraph{Discussion}
The results observed in Table \ref{tab:results-aggregation-new} highlights the versatility of the semantic segmentation model in trading-off true positives and true negatives; once the individual component networks are trained their outputs can be aggregated using different rules and threshold offering a further level of control to the user. A monolithic architecture made up by a single network does not have this option; while we could introduce a cost-sensitive loss function to trade-off precision and recall, any tuning of this loss function would requiring a re-training of the whole single network model.
However, the flexibility of the semantic segmentation model is earned at the cost of a hard combinatorial optimization problem; the new degrees of freedom mean that finding a combination of threshold and rule able to reach the raw level of accuracy of a single network model is not trivial.


\subsection{Simulation 4: Data Comparison of the Models}
Finally, we compare the performances of the two models in a low-data regime. In particular, we want to assess the hypothesis that the semantic decomposition model may be a better fit in low-data regimes since its task of detecting a simple component part may be arguably easier than the task of the single network model of detecting a whole weapon. It may be suggested that the task of a single network model is not more difficult when we take into account that such a model has the possibility of learning correlations between the component parts of a weapon; yet, we hypothesize that, in a low-data regime, data scarcity makes it challenging and unlikely for the single network model to learn such correlations, leaving the network with a harder task of detecting a weapon as a whole. In this section we put this hypothesis to test. This simulation again runs up to the \emph{network outputs level} of Figure \ref{fig:SSmodel}, like Simulation 1.

\paragraph{Protocol}
For this experiment we consider again the best performing architectures in term of accuracy that we discovered in the first simulation (see Table \ref{tab:perfromanceBest}).
Given that the performances on the whole dataset are known, we proceed to re-train and re-evaluate such architectures with random subsamples of the original dataset. We reduce the size of the training sets via random subsampling by considering only $75\%, 50\%,$ or $25\%$ or the original dataset, while keeping the size of the test dataset constant. We train and test following the same protocol of Simulation 1. 

\paragraph{Results}
Figure \ref{fig:limited-acc} shows how the final accuracy changes when the amount of training data shrinks (see Appendix for more details). Notice that, despite a small increase in performance of the single network model when trained on a 75\% limited dataset, the overall trend of this model presents a sharper drop than any of the networks in the semantic segmentation model.

\begin{figure}[h]
\centering
\caption{Results of limited training. \label{fig:limited-acc}}
\includegraphics[width=0.8\columnwidth]{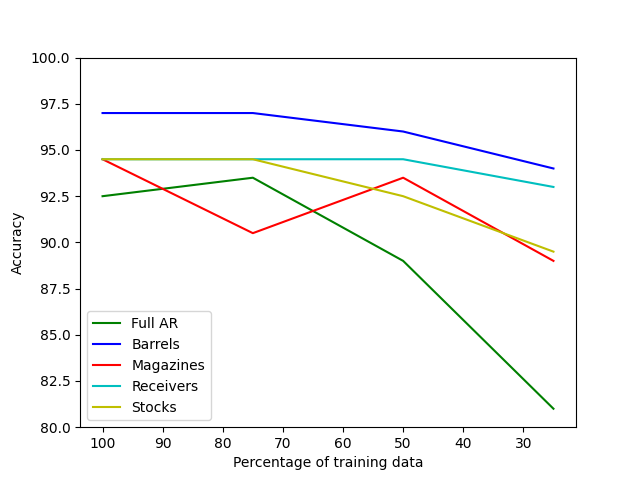}
\end{figure}

\paragraph{Discussion}
Smaller semantic networks proved to be more stable and accurate than the single network model when provided with a reduced set of data. This confirms the hypothesis that, in a low-data regime, without further changes to the architecture and the regularization, smaller solution may be a safer choice. \\

\section{Ethical considerations}\label{sec:EthicalConsiderations}
As in previous work \cite{egiazarov2020firearm}, we acknowledge that the application of machine learning model in critical scenarios presents potential ethical challenges. Our work is motivated by the development of system and tools that may benefit the civil society and that may be deployed to prevent violence and loss of lives. However, we are aware that a sensitive technology like weapon detection may find application in other contexts, for instance, within lethal autonomous weapon system (LAWS). As the authors of this work, we disavow such application of our work, and in particular we condemn the use of our models in autonomous weapons\footnote{\url{https://futureoflife.org/open-letter-autonomous-weapons/}}.

\section{Conclusions}

In this paper we considered two main approaches to the problem of detecting weapons in images: a standard, monolithic end-to-end approach based on a deep CNN, and an alternative modular approach based on the principle of decomposition of a complex problem in a set of smaller and simpler sub-problems. We conducted a set of rigorous experiments to evaluate the two solution from different points of view, within fixed computational limits.
Under the point of view of reliability and flexibility, the semantic segmentation model was shown to offer a higher degree of control to the designer: different sub-problems may be identified and solved by small dedicated CNNs, and the ratio between precision and recall may be controlled by changing the way the outputs of the individual component part networks are aggregated. This level of control is not available, by default, in a deep CNN, which automatically generates hierarchies of features, and which optimizes a loss function that does not explicitly account for precision and recall.
However, this added degree of freedom of the semantic segmentation model translates in a more challenging optimization problem. Thus, from the point of view of raw performance (in terms of accuracy, for instance) the single network model outperforms the semantic segmentation model thanks to its easier and more direct optimization, whereas the semantic segmentation model requires more fine-tuning of the aggregation parameters.
Yet, the semantic segmentation model proved to be more robust in lower-data regimes: decreasing the amount of available data has a smaller effect in the semantic segmentation model compared to the single network model, likely due to the fact that the individual component networks are learning simpler functions that can be fit with less data.
In summary, the semantic segmentation model was proven to show useful properties (flexibility, modularity, robustness) directly inherited from the underlying principle on which this model was designed. Its limited accuracy remains however a significant obstacle to make this model an alternative to the current deep CNN paradigm. Further work may aim at exploring more rigorous and grounded ways to deal with the problem of optimizing the aggregation process, either treating it as a hyper-parameter exploration problem or trying to solve it using black-box models.

\section*{Acknowledgment}
This research was supported by the research project Oslo Analytics
funded by the Research Council of Norway under the Grant No.: IKTPLUSS 247648.



\bibliographystyle{ieeetr}

\bibliography{references}
%

\newpage

\appendix

\setcounter{table}{0}
\renewcommand{\thetable}{A.\Roman{table}}

\subsection{Further experimental results \label{app:furtherResults}}
Tables \ref{tab:perfromanceBarrels}, \ref{tab:perfromanceMags}, \ref{tab:perfromanceReceivers}, and \ref{tab:perfromanceStocks} report the result of training the component networks of the semantic segmentation model on all the architectures we considered. Table \ref{tab:limit_training_res} reports the results of training both models on a limited data set.


\begin{table}[hbt!]
\caption{Best accuracy results for the network trained on the barrel component of the AR-15. \label{tab:perfromanceBarrels}}
\begin{tabular}{|c|c|c|c|c|}
\hline
\textbf{Arch} & \textbf{Epoch} & \textbf{Acc (val)} & \textbf{TP (test)} & \textbf{TN (test)} \\ \hline
\textbf{5x5}             & 15             & 94,4\%              & 97\%                  & 94\%                  \\ \hline
\textbf{4x4} & 6              & 95,2\%              & 99\%                  & 95\%                  \\ \hline
\textbf{4x3}       & 7              & 94,4\%              & 98\%                  & 96\%                  \\ \hline
\textbf{3x4}     & 12             & 92,5\%              & 92\%                  & 93\%                  \\ \hline
\textbf{3x3}             & 7              & 93,8\%              & 98\%                  & 91\%                  \\ \hline
\end{tabular}
\end{table}

\begin{table}[hbt!]
\caption{Best accuracy results for the network trained on the magazine component of the AR-15. \label{tab:perfromanceMags}}
\begin{tabular}{|c|c|c|c|c|}
\hline
\textbf{Arch} & \textbf{Epoch} & \textbf{Acc (val)} & \textbf{TP (test)} & \textbf{TN (test)} \\ \hline
\textbf{5x5}             & 14             & 90,7\%              & 92\%                  & 96\%                  \\ \hline
\textbf{4x4} & 11             & 93,0\%              & 93\%                  & 95\%                  \\ \hline
\textbf{4x3}       & 12             & 94,1\%              & 96\%                  & 93\%                  \\ \hline
\textbf{3x4}     & 7              & 91,5\%              & 85\%                  & 91\%                  \\ \hline
\textbf{3x3}             & 9              & 91,3\%              & 97\%                  & 92\%                  \\ \hline
\end{tabular}
\end{table}

\begin{table}[hbt!]
\caption{Best accuracy results for the network trained on the receiver component of the AR-15. \label{tab:perfromanceReceivers}}
\begin{tabular}{|c|c|c|c|c|}
\hline
\textbf{Arch} & \textbf{Epoch} & \textbf{Acc (val)} & \textbf{TP (test)} & \textbf{TN (test)} \\ \hline
\textbf{5x5}             & 8              & 97,7\%              & 99\%                  & 90\%                  \\ \hline
\textbf{4x4} & 11             & 96,6\%              & 99\%                  & 95\%                  \\ \hline
\textbf{4x3}       & 5              & 96,8\%              & 99\%                  & 92\%                  \\ \hline
\textbf{3x4}     & 5              & 95,6\%              & 98\%                  & 91\%                  \\ \hline
\textbf{3x3}             & 15             & 95,6\%              & 98\%                  & 90\%                  \\ \hline
\end{tabular}
\end{table}

\begin{table}[hbt!]
\caption{Best accuracy results for the network trained on the stock component of the AR-15. \label{tab:perfromanceStocks}}
\begin{tabular}{|c|c|c|c|c|}
\hline
\textbf{Arch} & \textbf{Epoch} & \textbf{Acc (val)} & \textbf{TP (test)} & \textbf{TN (test)} \\ \hline
\textbf{5x5}             & 11             & 94,8\%              & 98\%                  & 90\%                  \\ \hline
\textbf{4x4} & 10             & 93,9\%              & 97\%                  & 92\%                  \\ \hline
\textbf{4x3}       & 6              & 93,9\%              & 95\%                  & 92\%                  \\ \hline
\textbf{3x4}     & 5              & 91,1\%              & 94\%                  & 88\%                  \\ \hline
\textbf{3x3}             & 12             & 91,7\%              & 96\%                  & 88\%                  \\ \hline
\end{tabular}
\end{table}

\begin{table}[hbt!]
\caption{Accuracy of individual models after limited set training}
\label{tab:limit_training_res}
\begin{tabular}{lllll}
Training data                            & \textbf{25\%}               & \textbf{50\%}               & \textbf{75\%}               & \textbf{100\%}              \\ \hline
\multicolumn{1}{|l|}{\textbf{Full AR}}        & \multicolumn{1}{l|}{81\%}   & \multicolumn{1}{l|}{89\%}   & \multicolumn{1}{l|}{93.5\%} & \multicolumn{1}{l|}{92.5\%} \\ \hline
\multicolumn{1}{|l|}{\textbf{Barrels}}   & \multicolumn{1}{l|}{94\%}   & \multicolumn{1}{l|}{96\%}   & \multicolumn{1}{l|}{97\%}   & \multicolumn{1}{l|}{97\%}   \\ \hline
\multicolumn{1}{|l|}{\textbf{Magazines}} & \multicolumn{1}{l|}{89\%}   & \multicolumn{1}{l|}{93.5\%} & \multicolumn{1}{l|}{90.5\%} & \multicolumn{1}{l|}{94.5\%} \\ \hline
\multicolumn{1}{|l|}{\textbf{Receivers}} & \multicolumn{1}{l|}{93\%}   & \multicolumn{1}{l|}{94.5\%} & \multicolumn{1}{l|}{94.5\%} & \multicolumn{1}{l|}{94.5\%} \\ \hline
\multicolumn{1}{|l|}{\textbf{Stocks}}    & \multicolumn{1}{l|}{89.5\%} & \multicolumn{1}{l|}{92.5\%} & \multicolumn{1}{l|}{94.5\%} & \multicolumn{1}{l|}{94.5\%} \\ \hline
\end{tabular}
\end{table}

\end{document}